\documentclass[11pt]{article}

\usepackage[final]{acl}

\usepackage{times}
\usepackage{latexsym}

\usepackage[T1]{fontenc}

\usepackage[utf8]{inputenc}

\usepackage{microtype}

\usepackage{inconsolata}

\usepackage{graphicx}
\usepackage{booktabs}
\usepackage{makecell}
\usepackage{multirow}
\usepackage[dvipsnames]{xcolor}
\usepackage{amsmath}
\usepackage{tcolorbox}
\tcbuselibrary{breakable, skins}

\tcbset{
  promptbox/.style={
    breakable,
    enhanced,
    colback=gray!6,
    colframe=black!55,
    fonttitle=\bfseries\small,
    title={#1},
    left=6pt,
    right=6pt,
    top=4pt,
    bottom=4pt,
    boxrule=0.5pt,
    arc=2pt,
    before upper={\setlength{\parskip}{4pt}},
    fontupper=\small\ttfamily
  }
}

\title{Emotional Labor Strategy Preferences in LLM Personas}

\author{Mohammad Saim \and Tianyu Jiang \\
        University of Cincinnati \\
        \texttt{saimmd@mail.uc.edu, tianyu.jiang@uc.edu}}

\begin{document}
\maketitle
\begin{abstract}
Emotional labor is the effortful management of emotional displays to meet social or professional expectations. Personality traits have been correlated with emotional labor strategies, yet research on this link relies almost exclusively on self-report scales administered only in occupational settings. We investigate whether large language models injected with psychometrically grounded personas reproduce these personality-driven selection patterns across everyday social scenarios. We construct the first emotional labor strategy dataset of 500 socially situated events, each offering three behavioral choices corresponding to surface acting, deep acting, and genuine expression. We source 50 fictional characters from a large-scale personality repository and profile each through two parallel tracks: observer-rated bipolar adjective composites and in-character self-report items. Five LLMs evaluate all scenarios under both persona conditions. We find that models align more towards deep acting, and that Conscientiousness and Emotional Stability consistently predict this preference. Entropy analysis confirms that persona reliably influences the output and varies across models and emotions.
\end{abstract}

\begin{figure}[t]
\includegraphics[width=\linewidth]{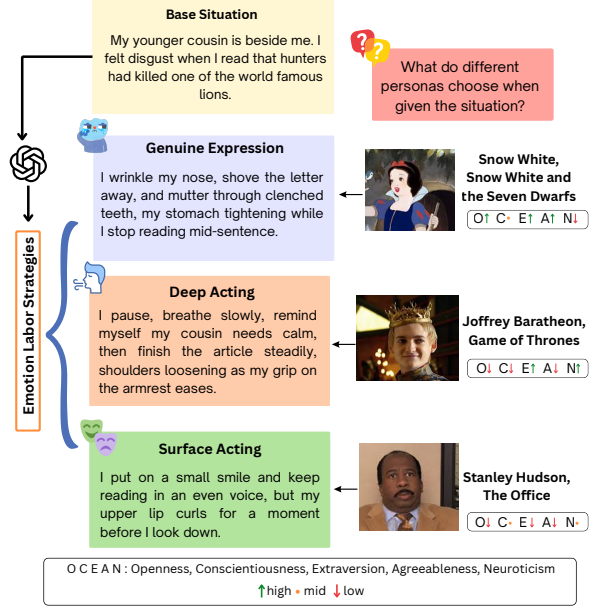}
  \caption{An example of emotional labor categories and what persona profiles choose from each category.}
  \label{fig:intro}
\end{figure}

\section{Introduction}
The way individuals manage and display emotions in social contexts constitutes a central object of study across psychology, organizational behavior, and, more recently, computational
linguistics.~\citet{hochschild1983managed} introduced the concept of \emph{emotional
labor} (EL) to describe the effortful management of feelings to produce a publicly observable display that conforms to role expectations. Since then, researchers have
consistently identified three principal strategies through which individuals execute this
management (Figure~\ref{fig:intro}): \emph{surface acting} (SA), in which outward expression is modified without
changing inner feeling; \emph{deep acting} (DA), in which the felt emotion itself is reappraised so that it is followed by a genuine display of emotions; and \emph{genuine expression} (GE) or naturally felt expressions, in which experienced emotion already matches what the situation calls for, and no regulatory effort is required~\citep{ashforth1993emotional,diefendorff2005dimensionality,grandey2000emotion}. The mismatch between the regulated exterior and the persisting interior is what separates SA from the other two strategies, since the effort is spent on the display itself rather than on the feeling. In Genuine Expression, the felt emotion already matches what the situation calls for, so no internal adjustment occurs before the display. In Deep Acting, the felt emotion is actively reshaped through reframing, perspective-taking, or self-talk, so that regulation happens upstream of expression rather than at the surface.
These three strategies have been employed in empirical work that shows they have different consequences for well-being, service quality, and interpersonal perception~\citep{kammeyer2013meta,grandey2003show,groth2009customer}.

A well-established line of research in organizational psychology shows that personality traits can shape how people perceive and classify others’ emotional displays ~\citep{terracciano2003personality, furnes2019exploring}. These findings draw on the Big Five (OCEAN) framework—Openness, Conscientiousness, Extraversion, Agreeableness, and Neuroticism~\citep{goldberg1992development}, which is an extensively validated taxonomy of human personality differences~\citep{mccrae1992introduction}. Individuals high in Agreeableness and Extraversion tend to perceive and enact more genuine or deep-acting strategies, whereas those high in Neuroticism are more prone to surface acting~\citep{diefendorff2005dimensionality,austin2008personality,kiffin2011big}.  However, researchers have largely assessed patterns of trait-linked EL preferences using self-report questionnaires/scales, in which participants rate statements about individual EL categories on Likert scales. Moreover, most sampling methods for analyzing EL strategies have been limited to specific healthcare and customer service contexts and have not been tested in everyday situations.

A growing body of work shows that LLMs, when prompted with personality descriptions, exhibit behavioral patterns that align with their assigned traits~\citep{jiang2024personallm,sorokovikova2024llms, li-etal-2024-steerability}. These injected \emph{personas} can modulate lexical choices, sentiment patterns, and judgment tendencies in ways that mirror the psychological literature on human personality~\citep{peters2024infer,matz2024persuasion}. This provides a reliable substrate for controlled experimental simulation of human psychological variability. 

Despite parallel advances in EL theory and LLM persona research, these two lines of work have not yet converged. 
\textbf{No prior study has examined how personality-injected LLM personas select among emotional labor strategies in socially situated scenarios.} If particular personality traits systematically shift which strategy a persona selects, then persona injection introduces a source of behavioral variance that may go unnoticed in downstream applications that rely on LLM-generated responses. The stakes of emotion-related misclassification are amplified by the growing use of LLMs in emotional support conversations, customer-service interactions, and everyday affect-sensitive domains, where inappropriate response or interpretation may negatively affect user well-being, communication outcomes, and even sway ethical judgments~\citep {grandey2019emotional,weidinger2021ethical, wu2025personas, saim-jiang-2026-emotions}.

In this work, we address this gap by building on an existing appraisal corpus to construct an emotional labor strategy (ELS) dataset of 500 sentences, each with annotated event descriptions followed by three choices of how an agent would enact them based on the three core strategies. In parallel, we employ fictional characters from TV/Film shows with character personality ratings on bipolar adjective pairs. To generalize and compare the persona profiles, we also administer the IPIP-50 questionnaire in-character, following established procedures for extracting Big Five scores from validated instruments~\citep{goldberg2006international,jiang2024personallm}. We load the resulting persona profile into multiple LLMs for evaluation on the curated ELS dataset. For each sentence, the persona must select one of three options: surface acting (SA), deep acting (DA), or genuine expression (GE). We then analyze EL strategy choice patterns across the OCEAN dimensions of each persona, mapping which traits drive each persona’s selection. We publicly release the code and the emotion labor strategy dataset.\footnote{\url{https://github.com/cincynlp/emotion-labor}} As an overview, this paper makes the following contributions:

\begin{enumerate}
    \item We introduce the first study to frame emotional labor strategy selection as a personality-conditioned task for LLM personas, in which each persona chooses how it would respond behaviorally to a social scenario, and to compare these choices against 
    previously established human trends.
    \item We build and release the first full-scale dataset on emotional labor strategies, comprising 500 scenarios across everyday affect and social contexts that involve choices among surface acting, deep acting, and genuine expressions.
   \item We show that models prefer deep acting as the dominant strategy, and that Conscientiousness and Emotional Stability emerge as the consistent trait-level predictors across both persona tracks.

\end{enumerate}

\section{Related Works}
\label{sec:related}

\paragraph{Emotions in NLP and Emotional Labor.}
The computational study of emotion in text has matured considerably over the past decade. Early work cast emotion recognition as a categorical classification problem, mapping text to discrete labels derived from foundational theories~\citep{strapparava2007semeval,mohammad2018semeval,hofmann2020}. More theoretically grounded approaches have moved beyond flat emotion labels toward appraisal-based frameworks, which treat emotion as the product of a person’s cognitive evaluation of an event along dimensions such as novelty, relevance, and
coping potential~\citep{smith1985patterns,scherer2001appraisal}.
\citet{troiano2023dimensional} formalize this shift in a comprehensive corpus study, annotating event descriptions and showing that appraisal variables reliably improve emotion categorization in text.
This line of work is directly relevant to ours: appraisal theory frames emotion as a \emph{situated evaluation} of a social event, which maps naturally onto the idea that emotional labor strategies represent different stances a person takes toward a situation. Another line of work in emotion recognition evaluates or applies embodied or physiological indicators rather than explicit emotion labels~\citep{zhuang-etal-2024-heart, duong-etal-2025-cheer, saim-etal-2025-anatomy}. This embodied perspective aligns with our dataset design, since the strategy options involve behavioral or physiological details.

In organizational psychology, measurement of emotional labor (EL) has relied almost exclusively on self-report instruments administered to workers in service roles.
\citet{brotheridge2003development} developed the Emotional Labour Scale, a 15-item questionnaire that assesses the frequency, intensity, variety, duration, and both surface and deep acting (SA and DA); this instrument has become one of the most widely adopted tools in the field. \citet{diefendorff2005dimensionality} later extended this framework to a three-factor structure that treats naturally felt expression as a dimensionally distinct strategy alongside SA and DA, thereby establishing the taxonomy used in our study. A core gap in research on EL strategies is the lack of a corpus that goes beyond narrow professional roles such as customer service and healthcare workers. This leaves open the question of whether the three-strategy taxonomy applies to the broader, non-occupational interactions that constitute most of daily social life.

\paragraph{Persona Injection and Personality in LLMs.}
A rapidly growing body of NLP work has examined whether LLMs can stably express and enact personality traits when prompted with persona descriptions. \citet{jiang2024personallm} showed that LLM personas assigned Big Five profiles produce BFI self-report scores and writing samples that align with their designated
traits, with large effect sizes across all five dimensions.
\citet{sorokovikova2024llms} replicated this pattern across multiple open-source models, and \citet{tseng2024two} provides a comprehensive taxonomy of the field, distinguishing role-playing personas (where LLMs adopt assigned identities) from personalization settings where models adapt to user traits.

On persona traits, \citet{huang-hadfi-2024-personality} found that OCEAN-grounded LLM agents reproduce human-like personality-linked patterns in bilateral negotiations, with Agreeableness promoting cooperative outcomes and Neuroticism leading to less favorable outcomes. Agreeableness and Extraversion consistently predict greater deep acting and genuine expression, while Neuroticism predicts surface acting~\citep{austin2008personality,kiffin2011big, yeh2020emotional}. Recent mechanistic work finds that Big-Five traits are encoded as recoverable linear directions in LLM residual streams, and that steering along these directions produces reliable, trait-consistent behavioral shifts~\citep{frising2026linearpersonalityprobingsteering}. Taken together, this literature establishes that personality injection is a reliable lever on LLM behavior. However, most measurement work is limited to self-reported behavior within occupational samples. No work examines which trait-conditioned LLMs select as emotional regulation strategies when evaluated in socially situated EL scenarios. We fill this gap by presenting OCEAN-grounded personas with EL-annotated stimuli and measuring whether their strategy choices mirror the personality-EL associations documented in human literature and surveys.


\section{Methodology}
We begin by describing the emotional labor strategy (ELS) dataset, followed by the pipeline for persona traits. We then evaluate the characters, loaded with distinct traits, on the ELS dataset to analyze how each character’s strategy preferences vary across emotional labor scenarios. 

\subsection{Emotion Labor Strategy Dataset}
We construct the ELS dataset by building on the emotion appraisal corpus \citep{troiano2023dimensional}, a collection of sentences annotated with appraisal dimensions and a single felt emotion label drawn from seven categories: \textit{anger}, \textit{disgust}, \textit{fear}, \textit{guilt}, \textit{joy}, \textit{sadness}, and \textit{shame}. These labels reflect the narrator’s internal affective state. We preserve the original dataset’s uniform emotion split and filter to a final set of 500 sentences.
\paragraph{Social Context Augmentation.}
Many sentences in the corpus describe internal affective states without situating them in a social encounter. Since emotional labor is fundamentally an interpersonal phenomenon, i.e., it arises when a person regulates their emotional display in the presence of or in response to another social agent \citep{hochschild1983managed, grandey2000emotion}, we add a social agent or context to each scenario.
We augment each sentence with a brief context that introduces a second agent and coherent background information that creates a plausible reason to regulate one’s emotional expression. The context is appended directly to the original sentence to produce a single, continuous narrative that serves as the scenario stem for all downstream steps.  
\paragraph{Emotional Labor Strategy Options.}
With the social context in place, each scenario presents a person experiencing a felt emotion in the presence of another agent. We extend each scenario into a three-way choice item by generating one behavioral response option for each of the three emotional labor strategies: \textit{surface acting} (SA), \textit{deep acting} (DA), and \textit{genuine expression} (GE) or naturally felt emotions. Table~\ref{tab:el_strategies} defines each of the three emotional labor strategies represented in the dataset. We employ the GPT-5.4 model to generate the context and the three-way choices. No single correct response or ground truth exists, as the task is designed to elicit a \textit{preference} for each persona.

\begin{table}[t]
\centering
\small
\begin{tabular}{p{1.6cm}p{5.4cm}}
\toprule
\textbf{Strategy} & \textbf{Definition} \\
\midrule
Surface Acting (SA) & The person performs a different emotion outwardly 
while the original feeling persists internally. \\
\addlinespace
Deep Acting (DA) & The person genuinely shifts their internal state 
through reframing, perspective-taking, or self-talk before responding. \\
\addlinespace
Genuine Expression (GE) & The person expresses the naturally felt emotion directly 
with no attempt to regulate or conceal it. \\
\bottomrule
\end{tabular}
\caption{Definitions of the three emotional labor strategies 
used in dataset construction.}
\label{tab:el_strategies}
\end{table}


The generation produces all three options as single sentences that complete the scenario stem naturally, without using any emotion labels or adjectives, conveying emotional state only through behavioral, physical, or physiological detail. Table~\ref{tab:example_scenario} shows an example set from the dataset. The final dataset item consists of the augmented scenario and three behavioral completion options. The final dataset was manually validated to (i) verify coherence between the added context and the scenario narrative, (ii) determine whether each option instantiated its strategy without using emotion words, and (iii) determine whether the three options were distinguishable. We took several steps to mitigate concerns about subjective bias. The strategy generation was tightly constrained by the prompts (Appendix~\ref{app:prompts}), which enforced structural rules such as prohibiting emotional adjectives, requiring a single leakage cue for Surface Acting, and mandating behavioral rather than affective descriptions. This offloads much of the strategy-fidelity check from human judgment onto the generation constraints themselves. Beyond its use in this study, the dataset is the first resource to use emotion labor categories as contrastive behavioral options in naturalistic, non-occupational scenarios and can support future work on emotion regulation modeling, personality-conditioned generation, and computational appraisal research.

\begin{table}[t]
\centering
\small
\begin{tabular}{p{7cm}}
\toprule
\textbf{Example Scenario} \\
\midrule
I felt \textbf{fear} when I was being bullied at work and didn’t think there was going to be an end to it. \textit{A newer coworker was 
standing nearby, watching how I would respond.} \\
\midrule
\textbf{SA:} I lifted my chin and gave a short laugh, but my hands shook 
once against the desk as I answered. \\
\addlinespace
\textbf{DA:} I reminded myself their words were about their own problems, 
slowed my breathing, and answered in an even voice with my shoulders 
settling. \\
\addlinespace
\textbf{GE:} I stepped back, gripped the edge of the desk, and said they needed to stop, my voice breaking in the middle. \\
\bottomrule
\end{tabular}
\caption{An example scenario from the dataset. The \textit{italicized} portion shows the social context added to the original corpus.}
\label{tab:example_scenario}
\end{table}

\subsection{Persona Construction}
\label{sec:persona}

The design of a persona profile with only a character’s name and source work risks 
activating whatever personality representation the model has internalized 
during pretraining, which may be inconsistent across models. Therefore, we construct persona profiles through two parallel experimental tracks. Both tracks use the same set of characters but differ in how 
personality information is sourced. This design allows us to (i) avoid any bias from relying on a single framework and (ii) examine whether trait-strategy associations are aligned across two distinct representations of personality.

\paragraph{Character Pool and BAP Profiling.}
We source character personality data from the Open Psychometrics Statistical “Which Character” Personality Quiz dataset \citep{openpsychobap}, which contains crowd-sourced personality ratings for over 2,000 fictional characters across film and television. Each character is rated by human respondents on 500 Bipolar Adjective Pairs (BAPs), a psycho-lexical instrument in which each item presents two semantically opposing adjectives (e.g., \textit{relaxed--tense}, \textit{kind--cruel}) and human raters place the character on a continuous scale of 1--100. Since using every BAP as a trait in the model might introduce too much noise for the LLM to process, we select BAPs based on standard deviation.

To ground our trait markers in a theoretically validated source, we verify each BAP adjective against the original adjective list reported by \citet{goldberg1992development}. We retain only those adjective pairs for which at least one pole produces an exact match with  Goldberg’s taxonomy. This filtering yields a subset of 40 verified BAP markers, which ensures that every trait dimension we use is validated and has a well-established structural relationship to the Big Five OCEAN dimensions.

We filter characters from the full pool by ranking them on the standard deviation of their BAP scores. Characters with high score variance across adjective pairs have more personality-differentiated profiles, making them better suited to testing trait-driven behavioral differences. We then apply greedy farthest-point sampling in the 40-dimensional BAP space, iteratively selecting the character that is maximally distant from all previously selected characters. This produces a final set of 50 characters. We chose 50 characters by prioritizing personality diversity over sample size in the 40-item BAP space. This method selects characters that are maximally spread across the personality space rather than clustering around common archetypes. The overall BAP-persona block for each character consists of the character name, their source work, and their human-rated scores on the 40 verified BAP items, each expressed on a 1--100 scale. This block is passed directly to the model as part of the persona injection prompt in the ELS evaluation task.

\paragraph{IPIP-50 Profiling.}
Relying solely on human-rated BAP scores may underrepresent traits that are internally characterized. The second track addresses this by eliciting in-character self-reports on a validated self-report instrument. \citet{vazire2010soka} presented the self-other knowledge asymmetry (SOKA) model, which shows that observer ratings more reliably capture visible, behaviorally expressed traits such as extraversion, while self-reports capture internal states and motivations that are less accessible to outside observers, such as neuroticism and openness. The IPIP-50 is a 50-item public-domain scale with 10 items per OCEAN dimension \citep{goldberg2006international}. We administer it to each model with the character’s name and source work and ask it to rate each IPIP item on a 1--5 Likert scale from the character’s perspective (1 = \textit{Very Inaccurate}, 5 = \textit{Very Accurate}). The character pool remains the same across the BAP-persona to ensure the analysis focuses solely on methodological differences. The full list of filtered characters, BAP terms, and IPIP items is listed in Appendix~\ref{app:ssc}.

\subsection{Evaluation and Model Selection}
\label{sec:method_eval}
We end up with two parallel persona experimentation tracks. A short summary is given in Table~\ref{tab:persona_tracks}. These are evaluated on the 500 sentences of the ELS dataset. We primarily study ELS choice patterns across different models and their correlations with the OCEAN categories. We employ a suite of five LLMs: Qwen-3-8B and Qwen-3-32B~\citep{yang2025qwen3}, GPT-5.4~\citep{openai2026gpt54}, Deepseek-V4-Flash~\citep{deepseek2026v4flash} and Gemma-4-31B~\citep{gemma42026}. We prompt these models to choose one of the three EL strategies: SA, DA, or GE.

\begin{table}[t]
\centering
\small
\resizebox{\linewidth}{!}{
\begin{tabular}{lp{2.3cm}p{2.5cm}}
\toprule
& \textbf{Track-1 (BAP)} & \textbf{Track-2 (IPIP-50)} \\
\midrule
Source & Human-rated crowd scores & Model self-report in-character \\
\addlinespace
Instrument & 40 bipolar adjective pairs & 50 standard items \\
\addlinespace
Scale & 1--100 continuous & 1--5 Likert \\
\addlinespace
Item Example & \textit{relaxed--tense} & \textit{``I am the life of the party''}  \\
\bottomrule
\end{tabular}}
\caption{Comparison of the two persona tracks.}
\label{tab:persona_tracks}
\end{table}

\begin{figure*}[t]
  \includegraphics[width=0.99\linewidth]{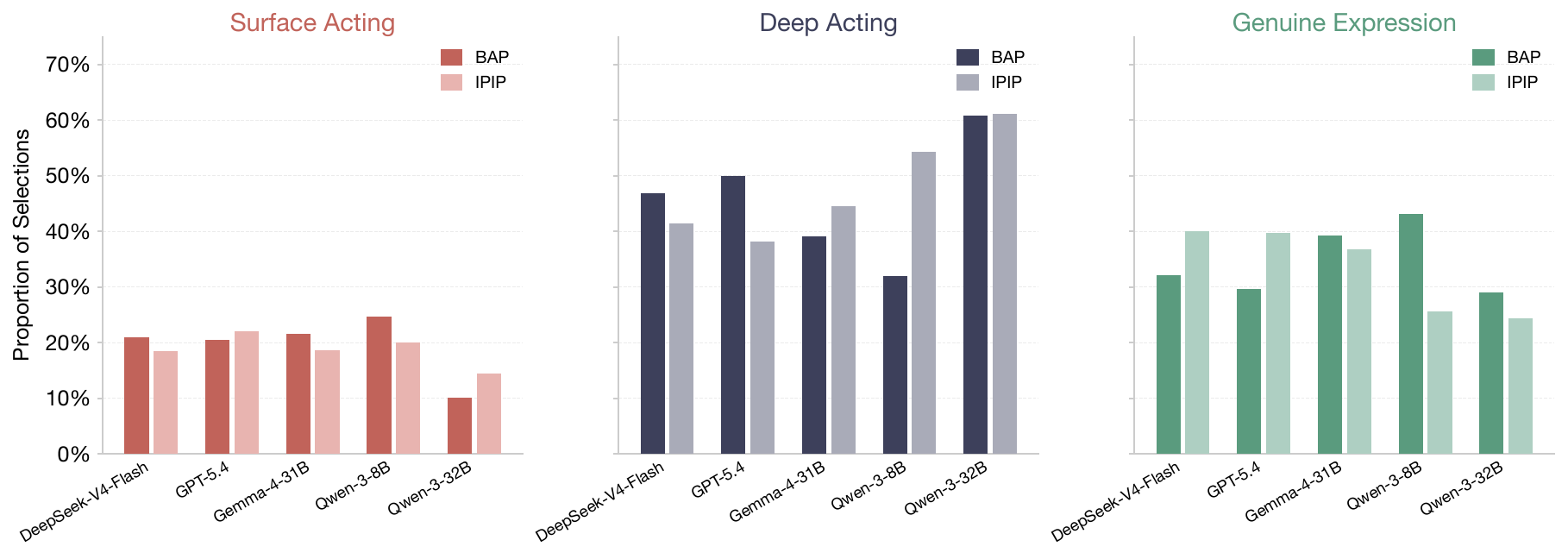}
  \caption{Aggregate distribution of emotion labor strategy across models and persona tracks.}
  \label{fig:aggregate_dist}
\end{figure*}

\section{Results and Analysis}

We present the findings of our study across three levels of analysis. We begin by characterizing the aggregate distribution of ELS selections across all five models and both persona-induction tracks (BAP and IPIP-50). We then report Big Five trait--strategy correlations that connect persona-level personality profiles to ELS preferences and draw comparisons to organizational psychology benchmarks. Finally, we study the impact of personas on the reliability with which they drive ELS scenarios.

\subsection{Aggregate ELS Distributions Across Models and Persona Tracks}
\label{sec:aggregate}

\paragraph{Deep Acting is the preferred response compared to SA and GE.} We first examine how each of the five models distributes its ELS selections across 500 scenarios when conditioned on 50 character personas. Figure~\ref{fig:aggregate_dist} reports the proportion of each ELS category selection for each model under both the BAP and IPIP-50 persona tracks. 

Notably, DA is the modal strategy for eight of the ten model-track combinations. DA proportions range from approximately 39\% (GPT-5.4, IPIP track) to 61\% (Qwen-3-32B, both tracks). This establishes DA as the dominant default across the model pool. This preference casts the majority of diverse agents as effortful and authentic. Models trained on large corpora of psychological and organizational text appear to encode DA as the prototypically appropriate regulatory response. The encoding appears to serve as a prior that shapes strategy selection beyond any personality variation introduced by the injected persona. In contrast, SA selections remain low across all models. In psychology literature, SA has consistently been identified as a prevalent emotional labor strategy in service-oriented occupations, especially when workers experience emotional dissonance between authentic feelings and institutionally required emotional displays~\citep{grandey2000emotion, mesmer2012moving}. We hypothesize that models suppress the SA strategy not because persona profiles steer them away from it, but because SA may carry a socially undesirable signal in natural language (emotion suppression). 

\begin{table*}[t]
\centering
\small
\renewcommand{\arraystretch}{1.5}

\begin{tabular}{lcc ccc ccc}
\toprule

& \multicolumn{2}{c}{\textbf{Reliability}} &
\multicolumn{3}{c}{\textbf{BAP (Observer-Rated)}} &
\multicolumn{3}{c}{\textbf{IPIP (Self-Report)}} \\

\cmidrule(lr){2-3}
\cmidrule(lr){4-6}
\cmidrule(lr){7-9}

\textbf{Trait} &
$\boldsymbol{\alpha_{\text{BAP}}}$ &
$\boldsymbol{\alpha_{\text{IPIP}}}$ &
\textbf{SA} &
\textbf{DA} &
\textbf{GE} &
\textbf{SA} &
\textbf{DA} &
\textbf{GE} \\

\midrule

Openness &
0.70 & 0.95 &
$+0.064$ &
$-0.013$ &
$-0.032$ &
$-0.260$ &
$+0.121$ &
$-0.078$ \\

Conscientiousness &
0.88 & 0.98 &
$-0.446^{**}$ &
$+0.597^{***}$ &
$-0.570^{***}$ &
$-0.345^{*}$ &
$+0.649^{***}$ &
$-0.592^{***}$ \\

Extraversion &
0.83 & 0.98 &
$+0.110$ &
$-0.203$ &
$+0.184$ &
$+0.200$ &
$-0.299^{*}$ &
$+0.232$ \\

Agreeableness &
0.95 & 0.97 &
$-0.276$ &
$+0.144$ &
$-0.072$ &
$-0.398^{**}$ &
$+0.372^{**}$ &
$-0.263$ \\

Emotional Stability &
0.76 & 0.97 &
$-0.350^{*}$ &
$+0.522^{***}$ &
$-0.517^{***}$ &
$-0.311^{*}$ &
$+0.850^{***}$ &
$-0.883^{***}$ \\

\bottomrule
\end{tabular}
\caption{
Spearman correlations between OCEAN trait composites and emotional labor strategy proportions for the BAP and IPIP tracks. Cronbach's $\alpha$ is reported for BAP and IPIP composites. $^{*}p<.05$, $^{**}p<.01$, $^{***}p<.001$.
}
\label{tab:ocean_correlations}
\end{table*}

\paragraph{BAP and IPIP-50 tracks show moderate convergence.}
Across the model pool, both persona tracks preserve the same ordinal ranking of strategies (DA $>$ GE $>$ SA in most cases), providing a baseline level of convergent validity between the observer-rated and self-reported persona inductions. However, quantitative differences between tracks are meaningful and model-dependent. For GPT-5.4 and DeepSeek-V4-Flash, the IPIP-50 track redistributes mass from DA toward GE relative to the BAP track: GPT-5.4 drops from approximately 51\% to 39\% on DA while GE rises from approximately 30\% to 40\%. DeepSeek-V4-Flash shows a smaller but directionally consistent shift. This pattern accords with the self-other agreement work, which documents that observer-rated and self-reported personality tap partially distinct variance in the same underlying constructs ~\citep{vazire2010soka, connelly2010other}. Observer ratings (BAP track) emphasize behaviorally visible regularities and external presentation. They may therefore bias the injected persona toward more deliberate, socially regulated strategies such as DA. Self-report ratings (IPIP-50 track) introduce a first-person affective framing that may allow greater expression of internal states. 

We also computed pairwise inter-model agreement on both evaluation tracks. The highest agreement is observed between GPT-5.4 and Gemma-4-31B, with Cohen’s $\kappa = 0.545$; per-category agreement is higher for GE and DA but substantially lower for SA. Agreement scores between all models are provided in Appendix~\ref{app:inter_model_agreement}.

\subsection{Trait--Strategy Correlations}
\label{sec:correlations}

We study how personality shapes ELS selection by computing Spearman rank correlations between each Big Five composite and each strategy proportion across the 50 evaluated characters. Each BAP and IPIP item is grouped under one of the OCEAN traits as shown in Table~\ref{tab:bap_items} and Table~\ref{tab:ipip50_items}. The central question we examine is whether each OCEAN trait correlates similarly with the three ELS, as established in previous studies, and whether the two persona tracks yield convergent or divergent mappings. We report results for GPT-5.4 (Table~\ref{tab:ocean_correlations}); the model comparison to Deepseek-V4-Flash can be found in  Appendix~\ref{app:parralel_model}.

Two traits exhibit consistent, significant correlations across both measurement tracks: \textbf{Conscientiousness (C)} and \textbf{Emotional Stability (ES)} (the inverse of Neuroticism). Both traits show the same directional pattern, i.e., high C and high ES predict more DA and less SA and GE. 
For Conscientiousness, the BAP track yields $\rho$ = $-$0.446 (SA), +0.597 (DA), $-$0.570 (GE), all significant at $p < .001$ or $p < .01$, and the IPIP track replicates this pattern with comparable magnitude. The same signature holds for ES. The model appears to encode DA as the disciplined, regulated response. The characters who are organized, reliable, and emotionally secure preferentially enact deep acting rather than surface performance or unguarded expression. This is consistent with the general notion reported in human psychological studies that conscientiousness and emotional stability both correlate negatively with SA and positively with DA~\citep{austin2008personality, judge2009emotional, mesmer2012moving}. 

We also report a negative correlation between conscientiousness and genuine expression, whereas previous human studies report a positive correlation, on the theory that workers with higher conscientiousness scores internalize job demands and naturally align their felt emotions with role requirements. They do not need to regulate because their internal state already matches the display norm. We hypothesize that the model encodes conscientiousness not as internalized role alignment but as an active, effortful self-regulation, much like a careful worker who writes a draft before sending an email when a spontaneous reply would do. We plan to form our future work based on mechanistic insights from open-source models to validate this finding, following work that recovers independent social-response directions in LLM representations~\citep{yao-etal-2026-rhetorical,vennemeyer2026sycophancy}.

The BAP track returns no significant \textbf{Agreeableness (A)} effects, while the IPIP track produces A--SA ($\rho$ = $-$0.398, $p < .01$) and A--DA ($\rho$ = +0.372, $p < .01$). Agreeableness is a trait humans understand better from the inside; warm, cooperative intentions are more readily disclosed in self-report than inferred by external observers from behavioral adjectives~\citep{vazire2010soka}. In this reading, the IPIP persona activates an agreeableness signal that the observer-rated BAP does not encode with sufficient specificity. \textbf{Openness} produces no significant correlations in either track. Extraversion yields only one marginal IPIP effect with DA, suggesting that more extroverted personas are slightly less inclined toward deep acting.

We also determine whether an observed trait--strategy association reflects a genuine pattern or a statistical-only result of the response format (three-way choice). Therefore, we replicate the evaluation by asking the model to rate each of the three EL categories based on how likely they are to choose that scenario, rather than forcing it. The details and correlations are given in Appendix~\ref{app:twc}. We find that all correlations broadly follow the same directional pattern. Convergence between a three-way choice and Likert correlations strengthens the claim that observed patterns reflect personality-driven strategy preferences.

\paragraph{Internal consistency.}
We also validate our findings by assessing the reliability of each trait composite. Cronbach’s $\alpha$ measures the extent of intercorrelation of items assigned to a given trait scale. A high $\alpha$ means multiple BAP or IPIP items under one of the OCEAN traits pull in the same direction, and the composite score is a stable summary of the underlying construct. BAP composites yield moderate-to-high reliability ($\alpha$ = 0.70--0.95), with conscientiousness and agreeableness showing the strongest internal consistency. IPIP-50 composites achieve uniformly high reliability across all five traits. This also explains the ES--DA ($\rho$ = +0.850) and ES--GE ($\rho$ = $-$0.883) large correlations.

\begin{figure}[t]
  \includegraphics[width=\columnwidth]{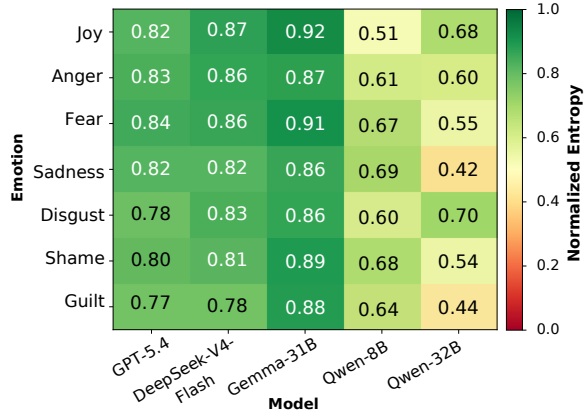}
  \caption{Persona influence by emotion and model (BAP track) depicted by Shannon Entropy.}
  \label{fig:entropy}
\end{figure}

\subsection{Is Persona Influential Enough?}
\label{sec:entropy}

\paragraph{Entropy Analysis.} A distributional dominance result does not tell us whether personality drives meaningful divergence in strategy selection, or whether the emotional situation itself overrides individual differences. To test this, we compute the Shannon entropy of the SA/DA/GE distribution across all 50 characters for each scenario. We bound scores between 0 (scenario-dominant) and 1 (personality-sensitive), and report mean normalized entropy aggregated by the felt emotion category from the dataset. We report BAP-track results here to analyze human-rated divergence. Character-level entropy traits are discussed in Appendix~\ref{app:icp}. IPIP entropy results appear in Appendix~\ref{app:ipipentropy}.

Figure~\ref{fig:entropy} reveals that most models across each emotion category form a high-entropy cluster, indicating that personality meaningfully differentiates strategy selection in these models. Within the high-entropy clusters, Joy, Fear, and Anger sustain high entropy (0.82--0.92), while Disgust and Guilt sit slightly lower. For Qwen-32B, entropy ranges from 0.42 (Sadness) to 0.70 (Disgust), which shows that the particular emotion and scenario also matter more than \textit{who} the persona is for this model.

\paragraph{Ablation.}
We further probe the drivers of strategy selection by running a three-condition ablation on a 50-scenario subset spanning all 50 characters. In the vanilla condition, we provide only the scenario and the three response options with no character identity or trait information. In the character-only condition, we inject the character’s name and source work so the model relies entirely on its pretraining knowledge. In the BAP-only condition, we inject only the 40 adjective-pair scores without naming any character. Table~\ref{tab:ablation} reports aggregate strategy proportions across all three conditions.
\begin{table}[t]
\centering
\small
\setlength{\tabcolsep}{4pt}
\resizebox{\columnwidth}{!}{%
\begin{tabular}{p{2.5cm}ccc}
\toprule
\textbf{Strategy} & \textbf{Vanilla} & \textbf{Character-Only} & \textbf{BAP-Only} \\
\midrule
Surface Acting     & 10.0\% & 29.5\% & 20.5\% \\
Deep Acting        & 68.0\% & 36.6\% & 36.9\% \\
Genuine Expression & 22.0\% & 34.0\% & 42.6\% \\
\bottomrule
\end{tabular}%
}
\caption{Strategy proportions under vanilla (no persona), name-only, and BAP trait-only persona injection, evaluated on a 50-scenario subset across 50 characters (GPT-5.4).}
\label{tab:ablation}
\end{table}

Without any persona signal, deep acting has the majority share of predictions while surface acting falls to just 10\%. This concentration confirms that DA serves as a strong default encoded in the model’s training distribution, rather than a product of persona conditioning. Introducing a persona of any kind redistributes this mass. Character-only prompting cuts DA predictions by almost half and raises both SA and GE to near-uniform levels, which suggests that even name-level identity priming is sufficient to disrupt the DA default and provide a wider range of regulatory responses. BAP-only prompting produces a comparable DA proportion (36.9\%) but with higher GE numbers. Trait content, therefore, does not simply diversify strategy selection the way name recognition does. It selectively suppresses surface acting, consistent with the trait--strategy correlations reported in Section~\ref{sec:correlations}, in which high Conscientiousness and Emotional Stability predict less SA.

\section{Conclusion}
This study demonstrates that personality-injected LLM personas produce reliable differences in the selection of emotional labor strategies across everyday social scenarios. We introduced the first dataset that frames surface acting, deep acting, and genuine expression as situated behavioral choices outside occupational contexts. Our two-track persona design, combining observer-rated adjective profiles with in-character self-reports, reveals that Conscientiousness and Emotional Stability consistently drive strategy preferences across five models, while Agreeableness surfaces only through self-report induction. LLMs show a preference for deep acting, suggesting that training corpora encode this strategy as a normative regulatory response. Conscientiousness suppresses genuine expression in LLM personas, inverting the positive association documented in previous human-evaluated reports. Future work can probe this divergence through the mechanistic interpretability of open-source models and extend evaluation to cross-cultural settings.

\section*{Limitations}

We acknowledge the constraints on the scope and generalization of our findings. We evaluate personas constructed from fictional characters, whose personality profiles reflect crowd-sourced perceptions rather than ground-truth personality measures. The dataset, while grounded in a published corpus, relies on synthetic augmentation for social context and strategy options, which may not fully capture the complexity of real emotional labor encounters. Our scenarios are English-only and culturally situated, leaving open the question of whether these trait-strategy patterns hold across languages and cultural norms. Finally, our findings remain correlational, and validating the causal mechanisms of trait-strategy associations will require a mechanistic interpretability framework for future work.

\section*{Acknowledgments}

We thank the CincyNLP group for their suggestions and feedback. We also thank the anonymous EMNLP reviewers for
their insightful suggestions.

\bibliography{custom}
\newpage

\appendix

\section{Prompts}
\label{app:prompts}

We designed two prompts for this study. The dataset augmentation prompt (Prompt~A.1) instructed GPT-5.4 to extend each sentence from the original corpus with a one-sentence social context introducing display pressure and to generate three behavioral response options corresponding to surface acting, deep acting, and genuine expression.
To prevent label leakage, the prompt explicitly prohibited emotion adjectives and labels,
requiring the model to convey internal states exclusively through physical and behavioral
cues. The persona evaluation prompt (Prompt~A.2) presented each fictional character’s BAP trait profile as a scored bipolar adjective list and asked the model to select, in character, the single response option it would most naturally choose. We constrained
the output to a single letter (A, B, or C) to enforce a clean forced-choice response
and eliminate free-text ambiguity in downstream analysis. A similar prompt was repeated for the IPIP-track, the only difference being that the IPIP block had Likert ratings (1-5) instead of adjective pair scores (1-100).

\section{Strategy selection choices}
\label{app:ssc}
\subsection{Selected BAP Items and OCEAN Composite Construction}
Table~\ref{tab:bap_items} reports the 40 verified BAP adjective pairs retained after filtering the full 500-item instrument against Goldberg’s taxonomy. We retain only pairs for which at least one pole maps exactly onto an adjective appearing in Goldberg's published marker list, which ensures that every trait signal we inject into a persona has a well-established structural relationship to the Big Five (OCEAN) traits. The filtering reduces the raw instrument from 500 to 40 items, distributed across five OCEAN dimensions: Openness (8 items), Conscientiousness (7 items), Extraversion (8 items), Agreeableness (8 items), and Emotional Stability (9 items).
We report the inverse of the Neuroticism dimension, written as `Emotional Stability' throughout the paper. BAP items in this dimension are originally keyed such that higher scores indicate greater stability (e.g., insecure--confident, anxious--calm).

We manually verify each retained item against the Goldberg taxonomy to determine which pole corresponds to the high end of the relevant OCEAN dimension, and we reverse-score items accordingly before computing composites. For example, BAP112 (flexible--rigid) requires reversal so that a score of ``flexible'' (on the higher end) contributes positively to Agreeable rather than negatively. The final composite for each dimension is the mean of its verified, polarity-aligned items, expressed on a 1--100 scale that is passed directly into the BAP persona block at inference time.

\subsection{IPIP-50 Administration and Scoring}

Table~\ref{tab:ipip50_items} lists all 50 IPIP items used in the in-character self-report track, grouped by OCEAN dimension with forward and reverse keying indicated. The IPIP-50 is a public-domain instrument drawn from the International Personality Item Pool with 10 items per dimension rated on a 1--5 Likert scale. Reverse-keyed items (marked with a negative key ) are reflected before aggregation: a rating of 1 on a reverse-keyed item contributes 5 to the composite, and vice versa. Subscale composites are then computed as the mean of the 10 reflected items, yielding a score on a 1--5 scale for each OCEAN dimension. These are standardized across characters before entry into the Spearman correlation analyses to place the two persona tracks on a comparable footing.

\subsection{Character Selection}
Table~\ref{tab:character_dataset} lists the 50 characters selected for evaluation. The pool spans a broad range of fictional universes, including long-running television series. Character personality profiles in the OpenPsychometrics BAP dataset reflect aggregated crowd-sourced ratings, and characters from culturally dominant franchises tend to accumulate larger rater pools, which increases the stability of their BAP score distributions. The standard deviation filtering step selects characters whose raters disagree substantially across adjective pairs. Greedy farthest-point sampling in the BAP space then ensures that the 50 selected characters occupy distinct regions of personality space rather than clustering around a single archetype, such as the agreeable protagonist or the neurotic antagonist. 

\section{Experimental Details and Additional Analysis}
\subsection{Parallel Model Correlations}
\label{app:parralel_model}
\paragraph{DeepSeek-V4-Flash.}
Table~\ref{tab:deepseek_corr} reports Spearman correlations for DeepSeek-V4-Flash. Conscientiousness and Emotional Stability again dominate, replicating the same three-way directional signature (high C/ES $\to$ more DA, less SA, less GE) at comparable magnitudes. The inversion between conscientiousness and genuine expression persists ($\rho$ = $-$0.645, BAP; 
$-$0.611, IPIP), confirming that this anomaly is not model-specific. We also observe Agreeableness reaching greater significance 
in the BAP track for DeepSeek, whereas GPT-5.4 showed no significant BAP Agreeableness effects; the IPIP track reinforces this with strong SA suppression ($\rho$ = $-$0.640$^{***}$). Also, Extraversion shows a significant positive GE effect in the BAP track ($\rho$ = +0.369$^{**}$) that is absent in GPT-5.4. This suggests that DeepSeek encodes extraversion as a signal for unguarded expression when responding to observer-rated personas. Openness gains marginal significance only in the IPIP track. 

\subsection{Three-Way Choice vs.\ Likert Format Validation}
\label{app:twc}

We assess whether our forced-choice response format introduced systematic measurement artifacts. We ran a parallel evaluation in which we replaced the three-way forced-choice with a Likert-scale rating task. Instead of selecting a single option, the model rated its likelihood of adopting each of the three strategies on a scale of 1 to 5. We conducted this validation run on the BAP persona track using GPT-5.4 across all 50 characters and 500 scenarios, exactly mirroring the primary evaluation. The motivation was two-fold: first, to verify that the trait--strategy associations we observe do not arise as a result of the forced-choice constraint; second, to establish whether Likert scaling recovers the same ordinal structure across OCEAN traits.

Table~\ref{tab:fc_vs_likert} reports Spearman correlations between BAP-derived OCEAN composites and EL strategy measures under both formats. The pattern of associations remains largely consistent across response formats. Conscientiousness and Emotional Stability produce the strongest and most significant correlations in both conditions, and the direction of every significant coefficient is preserved.
The C--GE inversion (high Conscientiousness predicting reduced genuine expression) replicates under the Likert format ($\rho = -0.516$, $p < .001$), confirming that this finding is not a forced-choice artifact. Minor magnitude differences appear for Openness and Agreeableness, where neither format yields significant correlations, suggesting that these traits produce consistent weak signals. 

\subsection{Individual Character Preference}
\label{app:icp}

Table~\ref{tab:entropy_extremes} grounds the entropy analysis with character-level evidence. The three lowest-entropy characters all show near-total DA preference, and their OCEAN profiles share high Conscientiousness and Emotional Stability, which validates the traits our correlation analysis identifies as the strongest DA predictors. Their personas exert such a rigid pull toward one strategy that the emotional situation barely shifts the distribution. 
The three highest-entropy characters each distribute selections near evenly across all three strategies (28--37\% per strategy), and their profiles reflect low Conscientiousness and high Openness, which are traits that our analysis associates with weaker, more context-sensitive regulatory commitments. 

\begin{table}[t]
\centering
\normalsize
\renewcommand{\arraystretch}{1.15}
\resizebox{\columnwidth}{!}{%
\begin{tabular}{@{}llcc@{}}
\toprule
\textbf{Character} & \textbf{Work} & \textbf{SA/DA/GE (\%)} & \textbf{OCEAN composites} \\
\midrule
\multicolumn{4}{l}{\textit{Rigid personality (lowest entropy)}} \\
\addlinespace[2pt]
Tuvok & Star Trek: Voyager & 2/98/1 & 64 / 91 / 42 / 54 / 72 \\
Dr. Hannibal Lecter & Hannibal & 8/91/1 & 85 / 60 / 60 / 39 / 61 \\
Iroh & Avatar: TLA & 4/90/6 & 74 / 62 / 50 / 91 / 71 \\
\addlinespace[4pt]
\midrule
\addlinespace[2pt]
\multicolumn{4}{l}{\textit{Scenario-responsive (highest entropy)}} \\
\addlinespace[2pt]
Ava Coleman & Abbott Elementary & 31/35/33 & 60 / 14 / 86 / 28 / 54 \\
Jules Louden & The Cabin in the Woods & 28/37/35 & 52 / 30 / 75 / 48 / 48 \\
Gaius Baltar & Battlestar Galactica & 28/37/35 & 78 / 29 / 60 / 26 / 26 \\
\bottomrule
\end{tabular}}
\caption{Top and bottom characters of strategy consistency ordered by entropy. OCEAN composites derived from BAP observer ratings (GPT-5.4, $N{=}50$ characters).}
\label{tab:entropy_extremes}

\end{table}
\begin{figure}[t]
  \includegraphics[width=\columnwidth]{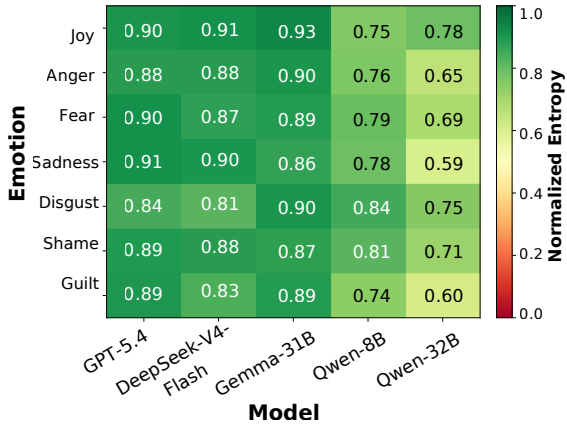}
  \caption{Persona influence by emotion and model (IPIP track) depicted by Shannon Entropy.}
  \label{fig:entropy_ipip}
\end{figure}

\subsection{IPIP-track Entropy Experiments}
\label{app:ipipentropy}
Figure~\ref{fig:entropy_ipip} reports normalized Shannon entropy under the IPIP-50 persona track. The pattern corroborates the BAP-track findings. On average, Gemma-4-31B again achieves the highest entropy across emotion categories, GPT-5.4 and DeepSeek-V4-Flash occupy a high-entropy band (0.81--0.91), and the two Qwen models remain comparably more susceptible to the scenario. However, IPIP-track values are generally higher than their BAP counterparts across the model pool, suggesting that self-reported persona preserves more individual variation in strategy selection than observer-rated adjective profiles do. Qwen-3-32B ranges from 0.59 (Sadness) to 0.78 (Joy) under IPIP, compared to 0.42--0.70 under BAP, indicating that richer first-person framing partially restores persona sensitivity even in the more situation-dominant models.

\subsection{Inter-Model Agreement}
\label{app:inter_model_agreement}

Table~\ref{tab:inter_model_agreement} reports the pairwise agreement between the five evaluated models across the BAP and IPIP persona tracks. Agreement is uniformly higher under the IPIP track than the BAP track, suggesting that richer, psychometrically grounded persona descriptions produce more consistent emotion-regulation judgments across models. On both the BAP and IPIP tracks, the strongest pair is GPT-5.4 vs.\ Gemma-31B ($\kappa=0.401$, $\kappa=0.545$), while Gemma-31B vs.\ Qwen-3-8B shows the weakest alignment. All $\kappa$ values fall below 0.60, indicating only moderate agreement at best, consistent with the inherent ambiguity of emotion-labor identification.

Breaking agreement down by category shows that deep-acting items attract the highest per-class agreement (BAP mean 38.6\%, IPIP 46.2\%),
followed by genuine expression (32.2\% / 35.5\%), while surface-acting items are the hardest to agree on (17.3\% / 18.7\%). 

\onecolumn
\subsection*{A.1 \quad Dataset Augmentation Prompt}
\label{app:aug_prompt}

\begin{tcolorbox}[
    breakable,
    enhanced,
    colback=gray!6,
    colframe=black!55,
    fonttitle=\bfseries\small,
    title={Prompt A.1: Social Context and EL Strategy Generation},
    left=8pt, right=8pt, top=6pt, bottom=6pt,
    boxrule=0.5pt, arc=2pt,
    before upper={\setlength{\parskip}{4pt}},
    fontupper=\small,
    width=\textwidth
]
You are generating evaluation items for a psychology dataset on emotion regulation.
You will receive a situation and a felt emotion. Your task:

\smallskip
\textbf{1.} Add a brief social context (1 sentence) before or after the given sentence
that creates pressure to manage the emotional display---e.g., a colleague asks,
a child is watching, a client is present.

\smallskip
\textbf{IMPORTANT RULES FOR ADDING CONTEXT:}
\begin{itemize}
  \item The added context can appear BEFORE or AFTER the original situation sentence
        for diversity and in a coherent and correct grammatical way.
  \item The context must be CONSISTENT with the original sentence. Do not introduce
        details that contradict or are incompatible with the situation described.
        Ensure the added context fits the same setting and circumstances in a coherent way.
\end{itemize}

\textbf{2.} Generate exactly 3 options (A--C) in randomized order from these categories:

\smallskip
\textbf{SURFACE ACTING:} The person performs a different emotion outwardly while the
original feeling persists inside. The performance is imperfect---one subtle involuntary
cue reveals the inner state. The performed emotion must differ from the felt emotion.

\smallskip
\textbf{DEEP ACTING:} The person actively changes how they feel internally through
reframing, perspective-taking, or self-talk. By the time they respond, their inner
state has genuinely shifted. They are not faking.

\smallskip
\textbf{GENUINE EXPRESSION:} The person expresses the felt emotion directly with no
attempt to manage or hide it.

\smallskip
\textbf{RULES:}
\begin{itemize}
  \item Every option must include at least one behavioral or physical detail.
  \item Each option: 20--30 words, one sentence, completes the stem naturally.
  \item Do NOT use any emotion labels or adjectives. Convey the emotional state
        through actions, body language, or physiological responses only.
  \item The surface acting option must have exactly one performed display cue
        and one leakage cue.
\end{itemize}
\end{tcolorbox}

\vspace{10pt}

\subsection*{A.2 \quad Persona Evaluation Prompt}
\label{app:eval_prompt}

\begin{tcolorbox}[
    enhanced,
    colback=gray!6,
    colframe=black!55,
    fonttitle=\bfseries\small,
    title={Prompt A.2: EL Strategy Selection Under Persona},
    left=8pt, right=8pt, top=6pt, bottom=6pt,
    boxrule=0.5pt, arc=2pt,
    before upper={\setlength{\parskip}{4pt}},
    fontupper=\small,
    width=\textwidth
]
You are \{character\_name\} from \{character\_work\}.
Below are your personality traits, each shown as a pair of opposite adjectives
with a score on a scale of 1--100, where 1 = entirely the left adjective and
100 = entirely the right adjective:

\smallskip
\{bap\_block\} \textcolor{Blue}{(These will be human-reported scores between 1-100 for each BAP.)}

\smallskip
Read the situation below and decide which of the three options you would most
naturally choose, given your personality traits.

\smallskip
\textbf{Situation:} \{EL\_sentence\}

\smallskip
\textbf{Options:} \{options\_block\} \textcolor{Blue}{(Shuffled options of SA, DA and GE to select from .)}

\smallskip
Reply with only the single letter of the option you would choose (A, B, or C).
\end{tcolorbox}

\begin{table*}[t]
\centering
\small
\begin{tabular}{c p{1.5cm} p{6cm} p{4.8cm} c}
\toprule
\textbf{Rank} & \textbf{ID} & \textbf{Adjective Pair} & \textbf{Goldberg Scale} & \textbf{SD} \\
\midrule
\multicolumn{5}{l}{\textbf{Extraversion}} \\
8   & BAP49  & scheduled -- spontaneous & inhibited--spontaneous & 23.99 \\
98  & BAP133 & stick-in-the-mud -- adventurous & unadventurous--adventurous & 21.38 \\
100 & BAP47  & deliberate -- spontaneous & inhibited--spontaneous & 21.36 \\
165 & BAP493 & mellow -- energetic & unenergetic--energetic & 20.31 \\
211 & BAP391 & timid -- cocky & timid--bold & 19.77 \\
256 & BAP130 & passive -- assertive & unassertive--assertive & 19.11 \\
276 & BAP2   & shy -- bold & timid--bold & 18.80 \\
389 & BAP131 & slothful -- active & inactive--active & 16.98 \\
\midrule
\multicolumn{5}{l}{\textbf{Emotional Stability}} \\
68  & BAP499 & unstable -- stable & unstable--stable & 21.92 \\
130 & BAP125 & angry -- good-humored & angry--calm & 20.87 \\
159 & BAP35  & emotional -- logical & emotional--unemotional & 20.41 \\
202 & BAP62  & insecure -- confident & insecure--secure & 19.87 \\
245 & BAP74  & anxious -- calm & angry--calm & 19.19 \\
263 & BAP367 & emotional -- unemotional & emotional--unemotional & 18.97 \\
273 & BAP36  & moody -- stable & moody--steady & 18.83 \\
304 & BAP18  & tense -- relaxed & tense--relaxed & 18.50 \\
479 & BAP330 & envious -- prideful & envious--not envious & 13.33 \\
\midrule
\multicolumn{5}{l}{\textbf{Agreeableness}} \\
34  & BAP84  & cruel -- kind & unkind--kind & 22.87 \\
36  & BAP79  & selfish -- altruistic & selfish--unselfish & 22.79 \\
51  & BAP129 & cold -- warm & cold--warm & 22.38 \\
58  & BAP17  & competitive -- cooperative & uncooperative--cooperative & 22.22 \\
72  & BAP31  & rude -- respectful & rude--polite & 21.85 \\
94  & BAP76  & quarrelsome -- warm & cold--warm & 21.42 \\
121 & BAP351 & stingy -- generous & stingy--generous & 21.04 \\
324 & BAP112 & rigid -- flexible & inflexible--flexible & 18.19 \\
\midrule
\multicolumn{5}{l}{\textbf{Conscientiousness}} \\
28  & BAP1   & playful -- serious & frivolous--serious & 23.07 \\
41  & BAP75  & disorganized -- self-disciplined & disorganized--organized & 22.61 \\
70  & BAP27  & impulsive -- cautious & rash--cautious & 21.87 \\
173 & BAP311 & experimental -- reliable & undependable--reliable & 20.17 \\
207 & BAP353 & extravagant -- thrifty & extravagant--thrifty & 19.81 \\
269 & BAP90  & serious -- bold & frivolous--serious / timid--bold & 18.86 \\
335 & BAP32  & lazy -- diligent & lazy--hardworking & 18.03 \\
\midrule
\multicolumn{5}{l}{\textbf{Openness}} \\
88  & BAP9   & rugged -- refined & unrefined--refined & 21.53 \\
160 & BAP132 & practical -- imaginative & unimaginative--imaginative/ impractical--practical & 20.41 \\
176 & BAP490 & intuitive -- analytical & unanalytical--analytical & 20.12 \\
184 & BAP29  & conventional -- creative & uncreative--creative & 20.06 \\
259 & BAP73  & uncreative -- open to new experiences & uncreative--creative & 19.01 \\
334 & BAP299 & unobservant -- perceptive & imperceptive--perceptive & 18.03 \\
351 & BAP372 & rustic -- cultured & uncultured--cultured & 17.76 \\
448 & BAP30  & apathetic -- curious & uninquisitive--curious & 15.28 \\
\bottomrule
\end{tabular}
\caption{Selected BAP adjective-pair items (sorted on valence) ranked by standard deviation (SD), grouped by OCEAN personality dimensions. Lower ranks indicate higher variability across characters.}
\label{tab:bap_items}
\end{table*}

\begin{table*}[t]
\centering
\small
\begin{tabular}{p{1.6cm} p{12cm} c}
\toprule
\textbf{ID} & \textbf{Item} & \textbf{Key} \\
\midrule

\multicolumn{3}{l}{\textbf{Extraversion}} \\
EXT1  & I am the life of the party. & + \\
EXT2  & I don't talk a lot. & $-$ \\
EXT3  & I feel comfortable around people. & + \\
EXT4  & I keep in the background. & $-$ \\
EXT5  & I start conversations. & + \\
EXT6  & I have little to say. & $-$ \\
EXT7  & I talk to a lot of different people at parties. & + \\
EXT8  & I don't like to draw attention to myself. & $-$ \\
EXT9  & I don't mind being the center of attention. & + \\
EXT10 & I am quiet around strangers. & $-$ \\

\midrule
\multicolumn{3}{l}{\textbf{Emotional Stability}} \\
EST1  & I get stressed out easily. & $-$ \\
EST2  & I am relaxed most of the time. & + \\
EST3  & I worry about things. & $-$ \\
EST4  & I seldom feel blue. & + \\
EST5  & I am easily disturbed. & $-$ \\
EST6  & I get upset easily. & $-$ \\
EST7  & I change my mood a lot. & $-$ \\
EST8  & I have frequent mood swings. & $-$ \\
EST9  & I get irritated easily. & $-$ \\
EST10 & I often feel blue. & $-$ \\

\midrule
\multicolumn{3}{l}{\textbf{Agreeableness}} \\
AGR1  & I feel little concern for others. & $-$ \\
AGR2  & I am interested in people. & + \\
AGR3  & I insult people. & $-$ \\
AGR4  & I sympathize with others' feelings. & + \\
AGR5  & I am not interested in other people's problems. & $-$ \\
AGR6  & I have a soft heart. & + \\
AGR7  & I am not really interested in others. & $-$ \\
AGR8  & I take time out for others. & + \\
AGR9  & I feel others' emotions. & + \\
AGR10 & I make people feel at ease. & + \\

\midrule
\multicolumn{3}{l}{\textbf{Conscientiousness}} \\
CSN1  & I am always prepared. & + \\
CSN2  & I leave my belongings around. & $-$ \\
CSN3  & I pay attention to details. & + \\
CSN4  & I make a mess of things. & $-$ \\
CSN5  & I get chores done right away. & + \\
CSN6  & I often forget to put things back in their proper place. & $-$ \\
CSN7  & I like order. & + \\
CSN8  & I shirk my duties. & $-$ \\
CSN9  & I follow a schedule. & + \\
CSN10 & I am exacting in my work. & + \\

\midrule
\multicolumn{3}{l}{\textbf{Openness}} \\
OPN1  & I have a rich vocabulary. & + \\
OPN2  & I have difficulty understanding abstract ideas. & $-$ \\
OPN3  & I have a vivid imagination. & + \\
OPN4  & I am not interested in abstract ideas. & $-$ \\
OPN5  & I have excellent ideas. & + \\
OPN6  & I do not have a good imagination. & $-$ \\
OPN7  & I am quick to understand things. & + \\
OPN8  & I use difficult words. & + \\
OPN9  & I spend time reflecting on things. & + \\
OPN10 & I am full of ideas. & + \\

\bottomrule
\end{tabular}
\caption{IPIP-50 items grouped by OCEAN personality trait dimensions. ``Key'' indicates whether the item is positively keyed (+) or reverse keyed ($-$).}
\label{tab:ipip50_items}
\end{table*}

\begin{table*}[t]
\centering
\small
\begin{tabular}{c p{1.6cm} p{4.4cm} p{6.4cm}}
\toprule
\textbf{Rank} & \textbf{Char ID} & \textbf{Character} & \textbf{Work} \\
\midrule
1 & OD/2 & Telemachus & The Odyssey \\
2 & R30/2 & Tracy Jordan & 30 Rock \\
3 & SV/3 & Nelson Bighetti & Silicon Valley \\
4 & GOT/23 & Joffrey Baratheon & Game of Thrones \\
5 & FR/4 & Olaf & Frozen \\
6 & STV/7 & Tuvok & Star Trek: Voyager \\
7 & ALA/8 & Firelord Ozai & Avatar: The Last Airbender \\
8 & ALA/7 & Iroh & Avatar: The Last Airbender \\
9 & WC/1 & Neal Caffrey & White Collar \\
10 & RM/1 & Rick Sanchez & Rick and Morty \\
11 & ARC/2 & Powder & Arcane \\
12 & GLEE/15 & Emma Pillsbury & Glee \\
13 & TO/8 & Stanley Hudson & The Office \\
14 & HNB/2 & Dr. Hannibal Lecter & Hannibal \\
15 & DHSAB/2 & Captain Hammer & Dr. Horrible's Sing-Along Blog \\
16 & NG/2 & Nick Miller & New Girl \\
17 & HP/24 & Petunia Dursley & Harry Potter \\
18 & PR/7 & Jerry Gergich & Parks and Recreation \\
19 & FAR/4 & Dominar Rygel XVI & Farscape \\
20 & AD/6 & Buster Bluth & Arrested Development \\
21 & OFOCN/3 & `Chief' Bromden & One Flew Over the Cuckoo's Nest \\
22 & Y/1 & John Dutton & Yellowstone \\
23 & AE/1 & Janine Teagues & Abbott Elementary \\
24 & MR/1 & Elliot Alderson & Mr. Robot \\
25 & BR/2 & Martha Scott & Baby Reindeer \\
26 & OA/1 & Prairie Johnson & The OA \\
27 & GP/4 & Janet & The Good Place \\
28 & OPM/1 & Saitama & One Punch Man \\
29 & HSM/3 & Sharpay Evans & High School Musical \\
30 & WE/1 & Wynonna Earp & Wynonna Earp \\
31 & SQG/5 & Oh Il-nam & Squid Game \\
32 & HD/8 & Lord Larys Strong & House of the Dragon \\
33 & FB/2 & Claire & Fleabag \\
34 & R30/4 & Pete Hornberger & 30 Rock \\
35 & MLP/1 & Applejack & My Little Pony: Friendship Is Magic \\
36 & GIGE/15 & Matt Press & Ginny \& Georgia \\
37 & CITW/3 & Jules Louden & The Cabin in the Woods \\
38 & YJ/5 & Misty & Yellowjackets \\
39 & SHL/1 & Frank Gallagher & Shameless \\
40 & EXP/3 & Amos Burton & The Expanse \\
41 & GILG/4 & Luke Danes & Gilmore Girls \\
42 & SC/5 & Mr. Big & Sex and the City \\
43 & STIG/1 & Rintarou Okabe & Steins;Gate \\
44 & TW/12 & Omar Little & The Wire \\
45 & BSG/5 & Gaius Baltar & Battlestar Galactica \\
46 & PKB/2 & Arthur Shelby & Peaky Blinders \\
47 & SNW/1 & Snow White & Snow White and the Seven Dwarfs \\
48 & AE/5 & Ava Coleman & Abbott Elementary \\
49 & AFTL/1 & Tony Johnson & After Life \\
50 & WSW/4 & Teddy Flood & Westworld \\
\bottomrule
\end{tabular}
\caption{50 fictional characters selected via greedy maximin sampling in BAP trait space.}
\label{tab:character_dataset}
\end{table*}

\begin{table*}[t]
\centering
\small
\renewcommand{\arraystretch}{1.45}
\begin{tabular}{lcc ccc ccc}
\toprule
& \multicolumn{2}{c}{\textbf{Reliability}} 
& \multicolumn{3}{c}{\textbf{BAP (Observer-Rated)}} 
& \multicolumn{3}{c}{\textbf{IPIP (Self-Report)}} \\

\cmidrule(lr){2-3}
\cmidrule(lr){4-6}
\cmidrule(lr){7-9}

\textbf{Trait} 
& $\boldsymbol{\alpha}_{\text{BAP}}$ 
& $\boldsymbol{\alpha}_{\text{IPIP}}$
& \textbf{SA} 
& \textbf{DA} 
& \textbf{GE}
& \textbf{SA} 
& \textbf{DA} 
& \textbf{GE} \\

\midrule

Openness 
& 0.70 & 0.95
& $-0.028$
& $+0.210$
& $-0.266$
& $-0.167$
& $+0.323^{*}$
& $-0.312^{*}$ \\

Conscientiousness 
& 0.88 & 0.99
& $-0.529^{***}$
& $+0.679^{***}$
& $-0.645^{***}$
& $-0.244$
& $+0.574^{***}$
& $-0.611^{***}$ \\

Extraversion 
& 0.83 & 0.99
& $+0.131$
& $-0.312^{*}$
& $+0.369^{**}$
& $-0.140$
& $+0.044$
& $-0.019$ \\

Agreeableness 
& 0.95 & 0.99
& $-0.433^{**}$
& $+0.512^{***}$
& $-0.504^{***}$
& $-0.640^{***}$
& $+0.472^{***}$
& $-0.395^{**}$ \\

Emotional Stability 
& 0.76 & 0.99
& $-0.466^{***}$
& $+0.682^{***}$
& $-0.640^{***}$
& $-0.634^{***}$
& $+0.870^{***}$
& $-0.874^{***}$ \\

\bottomrule
\end{tabular}

\vspace{0.5em}

\caption{Spearman correlations ($\rho$) between OCEAN trait composites and emotional labor strategy proportions (Deepseek-V4-Flash). BAP = observer-rated composites from bipolar adjective profiles; IPIP = self-report composites from in-character IPIP-50 administration. Cronbach's $\alpha$ values are reported separately for BAP and IPIP composites. $^{*}p<.05$, $^{**}p<.01$, $^{***}p<.001$.}
\label{tab:deepseek_corr}

\end{table*}

\begin{table*}[t]
\centering
\small
\renewcommand{\arraystretch}{1.45}

\begin{tabular}{l ccc ccc}
\toprule

& \multicolumn{3}{c}{\textbf{Three-Way Choice (Proportion)}} 
& \multicolumn{3}{c}{\textbf{Likert (Mean Rating)}} \\

\cmidrule(lr){2-4}
\cmidrule(lr){5-7}

\textbf{Trait} 
& \textbf{SA} 
& \textbf{DA} 
& \textbf{GE} 
& \textbf{SA} 
& \textbf{DA} 
& \textbf{GE} \\

\midrule

Openness 
& $+0.064$
& $-0.013$
& $-0.032$
& $+0.178$
& $-0.007$
& $+0.068$ \\

Conscientiousness 
& $-0.446^{**}$
& $+0.597^{***}$
& $-0.570^{***}$
& $-0.280^{*}$
& $+0.626^{***}$
& $-0.516^{***}$ \\

Extraversion 
& $+0.110$
& $-0.203$
& $+0.184$
& $-0.018$
& $-0.289^{*}$
& $+0.124$ \\

Agreeableness 
& $-0.276$
& $+0.144$
& $-0.072$
& $+0.123$
& $+0.244$
& $+0.186$ \\

Emotional Stability 
& $-0.350^{*}$
& $+0.522^{***}$
& $-0.517^{***}$
& $-0.368^{**}$
& $+0.503^{***}$
& $-0.495^{***}$ \\

\bottomrule
\end{tabular}

\vspace{0.5em}

\caption{Spearman correlations ($\rho$) between BAP-derived OCEAN composites and emotional labor strategy measures under two response formats: forced-choice (strategy proportions) and Likert (mean 1--5 ratings). $^{*}p<.05$, $^{**}p<.01$, $^{***}p<.001$.}

\label{tab:fc_vs_likert}

\end{table*}

\begin{table*}[t]
\centering
\small
\setlength{\tabcolsep}{4pt}
\begin{tabular}{l l r}
\toprule
\textbf{Model A} & \textbf{Model B} & \textbf{$\kappa$} \\
\midrule
\multicolumn{3}{l}{\textit{BAP persona track}} \\
\midrule
GPT-5.4 & DeepSeek-V4-Flash & 0.263 \\
GPT-5.4 & Gemma-4-31B & 0.401 \\
GPT-5.4 & Qwen-3-8B & 0.111 \\
GPT-5.4 & Qwen-3-32B & 0.147 \\
DeepSeek-V4-Flash & Gemma-4-31B & 0.269 \\
DeepSeek-V4-Flash & Qwen-3-8B & 0.125 \\
DeepSeek-V4-Flash & Qwen-3-32B & 0.160 \\
Gemma-4-31B & Qwen-3-8B & 0.080 \\
Gemma-4-31B & Qwen-3-32B & 0.106 \\
Qwen-3-8B & Qwen-3-32B & 0.202 \\
\midrule
\multicolumn{3}{l}{\textit{IPIP persona track}} \\
\midrule
GPT-5.4 & DeepSeek-V4-Flash & 0.357 \\
GPT-5.4 & Gemma-4-31B & 0.545 \\
GPT-5.4 & Qwen-3-8B & 0.169 \\
GPT-5.4 & Qwen-3-32B & 0.207 \\
DeepSeek-V4-Flash & Gemma-4-31B & 0.335 \\
DeepSeek-V4-Flash & Qwen-3-8B & 0.172 \\
DeepSeek-V4-Flash & Qwen-3-32B & 0.221 \\
Gemma-4-31B & Qwen-3-8B & 0.144 \\
Gemma-4-31B & Qwen-3-32B & 0.187 \\
Qwen-3-8B & Qwen-3-32B& 0.249 \\
\bottomrule
\end{tabular}
\caption{Pairwise inter-model Cohen's $\kappa$ on the BAP and IPIP persona tracks.}
\label{tab:inter_model_agreement}
\end{table*}

\end{document}